\pdfoutput=1

\documentclass[11pt]{article}

\usepackage{ACL2023}

\usepackage{times}
\usepackage{latexsym}

\usepackage[T1]{fontenc}

\usepackage[utf8]{inputenc}

\usepackage{microtype}

\usepackage{inconsolata}

\def\ccp#1{%
    \pgfmathsetmacro\calc{(#1)*100/(100)}%
    \edef\clrmacro{\noexpand\cellcolor{blue!\calc}}%
    \clrmacro%
    \ifdim \calc pt>50pt\color{white}\fi{#1}%
}

\usepackage{multirow}
\usepackage{lipsum} 
\usepackage{adjustbox}
\usepackage{comment}
\usepackage{xurl}
 \usepackage{float}
\usepackage{amssymb}

\usepackage{xcolor}

\title{Robust Hate Speech Detection in Social Media: \\ A Cross-Dataset Empirical Evaluation}

\author{Dimosthenis Antypas \and Jose Camacho-Collados \\
        Cardiff NLP, School of Computer Science and Informatics \\ Cardiff University, United Kingdom \\  \texttt{\{AntypasD,CamachoColladosJ\}@cardiff.ac.uk} \\}

\begin{document}
\maketitle
\begin{abstract}

The automatic detection of hate speech online is an active research area in NLP. Most of the studies to date are based on social media datasets that contribute to the creation of hate speech detection models trained on them. However, data creation processes contain their own biases, and models inherently learn from these dataset-specific biases. In this paper, we perform a large-scale cross-dataset comparison where we fine-tune language models on different hate speech detection datasets. This analysis shows how some datasets are more generalisable than others when used as training data. Crucially, our experiments show how combining hate speech detection datasets can contribute to the development of robust hate speech detection models. This robustness holds even when controlling by data size and compared with the best individual datasets. 

\end{abstract}

\section{Introduction}

Social media has led to a new form of communication that has changed how people interact across the world. With the emergence of this medium, hateful conduct has also found a place to propagate online. From more obscure online communities such as 4chan \cite{knuttila2011user} and Telegram rooms \cite{walther2021us} to mainstream social media platforms such as Facebook \cite{del2017hate} and Twitter \cite{udanor2019combating}, the spread of hate speech is an on going issue. 

Hate speech detection is a complex problem that has received a lot of attention from the Natural Language Processing (NLP) community. It shares a lot of challenges with other social media problems (emotion detection, offensive language detection, etc),  such as an increasingly amount of user generated content, unstructured  \cite{elsayed2019proposed} and constantly evolving text \cite{ebadi2021understanding}, and the need of efficient large scale solutions. When dealing with hate speech in particular, one has to consider the sensitivity of the topics, their wide range (e.g. sexism, sexual orientation, racism), and their evolution through time and location \cite{matamoros2021racism}. Understanding the extent of the problem and tracking hate speech online through automatic techniques can therefore be part of the solution of this ongoing challenge. One way to contribute to this goal is to both improve the current hate speech detection models and, crucially, the data used to train them.

The contributions of this paper are twofold. First, we provide a summary and unify existing hate speech detection datasets from social media, in particular Twitter. Second, we analyse the performance of language models trained on all datasets, and highlight deficiencies in generalisation across datasets, including the evaluation in a new independently-constructed dataset. Finally, as a practical added value stemming from this paper, we share all the best models trained on the unification of all datasets, providing a relatively small-size hate speech detection model that is generalisable across datasets.\footnote{The best binary hate speech detection model is available at \url{https://huggingface.co/cardiffnlp/twitter-roberta-base-hate-latest}; the multiclass hate speech detection model identifying target groups is available at \url{https://huggingface.co/cardiffnlp/twitter-roberta-base-hate-multiclass-latest}. These models have been integrated into the TweetNLP library \cite{camacho-collados-etal-2022-tweetnlp}.}

\paragraph{Content Warning} The article contains examples of hateful and abusive language. The first vowel in hateful slurs, vulgar words, and in general profanity language is replaced with an asterisk (*).

\section{Related Work}
Identifying hate speech in social media is an increasingly important research topic in NLP. It is often framed as a classification task (binary or multiclass) and through the years various machine learning and information sources approaches have been utilised \cite{mullah2021advances,ali2022hate,khanday2022detecting,del2023socialhaterbert}. A common issue of supervised approaches lies not necessarily with their architecture, but with the existing hate speech datasets that are available to train supervised models. It is often the case that the datasets are focused on specific target groups \cite{grimminger-klinger-2021-hate}, constructed using some specific keyword search terms \cite{waseem-hovy-2016-hateful,zampieri-etal-2019-predicting}, or have particular class distributions \cite{basile-etal-2019-semeval} that leads to a training process that may or may not generalise. For instance, \citet{florio2020time} analysed the temporal aspect of hate speech, and demonstrate how brittle hate speech models are when evaluated on different periods. Recent work  has also shown that there is a need to both focus on the resources available and also try to expand them in order to develop robust hate speech classifiers that can be applied in various context and in different time periods \cite{bourgeade-etal-2023-learn, bose-etal-2022-dynamically}.

In this paper, we perform a large-scale evaluation to analyse how generalisable supervised models are depending on the underlying training set. Then, we propose to mitigate the relative lack of generalisation by using datasets from various sources and time periods aiming to offer a more robust solution. 



\section{Data}

In this section, we describe the data used in our experiments. First, we describe existing hate speech datasets in Section \ref{hatedatasets}. Then, we unify those datasets and provide statistics of the final data in Section \ref{unification}

    \subsection{Hate Speech datasets}
    \label{hatedatasets}
    In total, we collected 13 datasets related to hate speech in social media. The datasets selected are diverse both in content, different kind of hate speech, and in a temporal aspect.
        
    \paragraph{Measuring hate speech (MHS)}
    \textit{MHS} \cite{kennedy2020constructing, sachdeva-etal-2022-measuring}  consists of 39,565  social media (YouTube, Reddit, Twitter) manually annotated comments. The coders were asked to annotate each entry on 10 different attributes such as the presence of sentiment, respect, insults and others; and also indicate the target of the comment (e.g. age, disability). They use Rasch measurment theory \cite{rasch1960studies} to aggregate the annotators' rating in a continuous value that indicates the hate score of the comment. 
    
    \paragraph{Call me sexist, but (CMS)} 
    This dataset of 6,325 entries \cite{samory2021call} focuses on the aspect of sexism and includes social psychology scales and tweets extracted by utilising the "Call me sexist, but" phrase. The authors also include two other sexism datasets \cite{jha-mamidi-2017-compliment, waseem-hovy-2016-hateful} which they re-annotate. Each entry is annotated by five coders and is labelled based on its content (e.g. sexist, maybe-sexist) and phrasing (e.g. civil, uncivil).
    
    \paragraph{Hate Towards the Political Opponent (HTPO)}
    \textit{HTPO} \cite{grimminger-klinger-2021-hate} is a collection of 3,000 tweets related to the 2020 USA presidential election. The tweets were extracted using a set of keywords linked to the presidential and vice presidential candidates and each tweet is annotated for stance detection (in favor of/against the candidate) and whether it contains hateful language or not.
    
    \paragraph{HateX}
    \textit{HateX} \cite{mathew2021hatexplain} is a collection of 20,148 posts from Twitter and Gab extracted by utilising relevant hate lexicons. For each entry, three annotators are asked to indicate: (1) the existence of hate speech, offensive speech, or neither of them, (2) the target group of the post (e.g. Arab, Homosexual), and (3) the reasons for the label assigned. 
    
    \paragraph{Offense}
    The \textit{Offense} dataset \cite{zampieri-etal-2019-predicting} contains 14,100 tweets extracted by utilising a set of keywords and categorises them in three levels: (1) offensive and non-offensive; (2) targeted/untargeted insult; (3) targeted to individual, group, or other. 
    
    \paragraph{Automated Hate Speech Detection (AHSD)}
    In this dataset, \cite{davidson2017automated} the authors utilise a set of keywords to extract 24,783 tweets which are manually labelled as either hate speech, offensive but not hate speech, or neither offensive nor hate speech. 
    
    \paragraph{Hateful Symbols or Hateful People? (HSHP)}
    This is a collection \cite{waseem-hovy-2016-hateful} of 16,000 tweets extracted based on keywords related to sexism and racism. The tweets are annotated as on whether they contain racism, sexism or neither of them by three different annotators.\footnote{A subset of the dataset is included in the \textit{Call me sexist, but} and is not considered.}

    \paragraph{Are You a Racist or Am I Seeing Things? (AYR)}
    This dataset \cite{waseem-2016-racist} is an extension of \textit{Hateful Symbols or Hateful People?} and adds the "both" (sexism and racism) as a potential label. Overlapping tweets were not considered.

    \paragraph{Multilingual and Multi-Aspect Hate Speech Analysis (MMHS)}
    MMHS \cite{ousidhoum-etal-2019-multilingual} contains hateful tweets in three different languages (English, French, Arabic). Each tweet has been labelled by three annotators on five different levels: (1) directness, (2) hostility (e.g. abusive, hateful), (3) target (e.g. origin, gender), (4) group (e.g. women, individual) and (5) annotator emotion (disgust, shock, etc). A total of 5,647 tweets are included in the dataset.

    \paragraph{HatE}
    \textit{HatE} \cite{basile-etal-2019-semeval} consists of English and Spanish tweets (19,600 in total) that are labelled on whether they contain hate speech or not. The tweets in this dataset focus on hate speech towards two groups: (1) immigrants and (2) women.

    \paragraph{HASOC}
    This dataset  \cite{mandl2019overview} contains 17,657 tweets in Hindi, German and English which are annotated on three levels: (1) whether they contain hate-offensive content or not; (2) in the case of hate-offensive tweets, whether a post contains hate, offensive, or profane content/words; (3) on the nature of the insult (targeted or un-targeted).

    \paragraph{Detecting East Asian Prejudice on Social Media (DEAP)}
    This is a collection of 20,000 tweets \cite{vidgen-etal-2020-detecting} focused on East Asian prejudice, e.g. Sinophobia, in relation to the COVID-19 pandemic. The annotators were asked to labelled each entry based on five different categories (hostility, criticism, counter speech, discussion, non-related) and also indicate the target of the entry (e.g. Hong Kongers, China).
   
    \paragraph{Large Scale Crowdsourcing and Characterization of Twitter Abusive Behavior (LSC)}
    The dataset \cite{founta2018large} consists of 80,000 tweets extracted using a boosted random sample technique. Each tweet is labelled as either offensive, abusive, hateful, aggressive, cyberbullying or normal.

    \subsection{Unification}
    Even though all of the datasets that were collected revolve around hate speech, there are major differences among them in terms of both format and content. We attempt to unify the datasets by standarizing their format and combining the available content into two settings: (1) binary hate speech classification and (2) a multiclass classification task including the target group.  We note that in cases where the original annotation results were provided, we decided to assign a label if at least two of the coders agree on it and not necessarily the majority of them. This approach can lead to a more realistic dataset and contribute in creating more robust systems \cite{antypas-etal-2022-twitter, mohammad2018semeval}.
    \label{unification}
        \subsubsection{Initial preprocessing}
        
        For each dataset collected, a simple preprocessing pipeline is applied. Firstly, any non-Twitter content is removed; despite the similarities between the content shared in various social media (e.g. internet slang, emojis), Twitter displays unique characteristics, such as the concept of retweets and shorter texts, which differentiate it from other platforms such as Reddit or Youtube \cite{smith2012does}. 
        Moreover, as our main consideration is hate speech in the English language, we exclude any non-English subset of tweets, and also verify the language by using a fastText based language identifier \cite{bojanowski-etal-2017-enriching}. Finally, considering that some datasets in this study utilise similar keywords to extract tweets, we remove near duplicated entries to avoid any overlap between them. This is accomplished by applying a normalisation step where entries that are considered duplicated based on their lemmatised form are ignored. Also, all URLs and mentions are removed.

        As a final note, three of the datasets (\textit{HSHP, AYR, LSC}) were dehydrated using the Twitter API since only their tweet IDs and their labels were publicly available. Unfortunately, a significant number of tweets ($\approx10,000$) were no longer available from the API.

        \subsubsection{Binary Setting}
        The majority of the datasets collected are either set as a binary hate classification task and no further preprocessing is applied (\textit{HTPO}), or offer a more fine-grained classification of hate speech (e.g. \textit{HateX}, \textit{CMS}) where we consider all "hate" subclasses as one. In general, when a dataset focuses on a specific type of hate speech (e.g. sexism) we map it as hate speech. Notable exceptions are: (1) The \textit{MSH} dataset, where a continues hate score is provided which is  transformed into a binary class according to the mapping proposed by the original authors. (2) Datasets that consist of offensive speech but also provide information about the target of the tweet. In these cases, (\textit{Offense}), we consider only entries that are classified as offensive and are targeting a group of people and not individuals. Our assumption is that offensive language towards a group of people is highly likely to target protected characteristics and thus be classified as hate speech. (3) Finally, only entries classified as hate speech were considered in datasets where there is a clear distinction between hate, offensive, or profound speech (\textit{LSC},  \textit{AHSD}, \textit{HASOC}). All data labelled as normal or not-hateful are also included as \textit{not-hate} speech.

\begin{table*}[ht]
\scalebox{0.95}{
\begin{tabular}{|l|cc|ccclll|}
\hline
\multicolumn{1}{|c|}{\multirow{2}{*}{\textbf{Dataset}}} & \multicolumn{2}{c|}{\textbf{Binary}}                     & \multicolumn{6}{c|}{\textbf{Multiclass}}                                                                                                                                                                                                                  \\ \cline{2-9} 
\multicolumn{1}{|c|}{}                                  & \multicolumn{1}{c|}{\textbf{hate}}   & \textbf{not-hate} & \multicolumn{1}{c|}{\textbf{racism}} & \multicolumn{1}{c|}{\textbf{sexism}} & \multicolumn{1}{c|}{\textbf{sexual orientation}} & \multicolumn{1}{c|}{\textbf{disability}} & \multicolumn{1}{c|}{\textbf{religion}} & \multicolumn{1}{c|}{\textbf{other}} \\ \hline
HatE                                                       & \multicolumn{1}{c|}{5303}            & 7364              & \multicolumn{1}{c|}{2474}            & \multicolumn{1}{c|}{2829}            & \multicolumn{4}{c|}{-}                                                                                                                                                      \\ \hline
MHS                                                       & \multicolumn{1}{c|}{2485}            &    5074           & \multicolumn{1}{c|}{735}             & \multicolumn{1}{c|}{784}             & \multicolumn{1}{c|}{251}                          & \multicolumn{1}{l|}{21}                  & \multicolumn{1}{l|}{246}               & 10                                  \\ \hline
DEAP                                                       & \multicolumn{1}{c|}{3727}            & 105               & \multicolumn{1}{c|}{3727}            & \multicolumn{5}{c|}{-}                                                                                                                                                                                             \\ \hline
CMS                                                       & \multicolumn{1}{c|}{1237}            & 10861             & \multicolumn{1}{c|}{-}               & \multicolumn{1}{c|}{1237}            & \multicolumn{4}{c|}{-}                                                                                                                                                      \\ \hline
Offense                                                       & \multicolumn{1}{c|}{1142}            & 12547             & \multicolumn{6}{c|}{-}                                                                                                                                                                                                                                    \\ \hline
HateX                                                       & \multicolumn{1}{c|}{2562}            & 5678              & \multicolumn{1}{c|}{757}             & \multicolumn{1}{c|}{492}             & \multicolumn{1}{c|}{407}                          & \multicolumn{1}{l|}{30}                  & \multicolumn{1}{l|}{239}               & 143                                 \\ \hline
LSC                                                       & \multicolumn{1}{c|}{889}             & 1267              & \multicolumn{6}{c|}{-}                                                                                                                                                                                                                                    \\ \hline
MMHS                                                       & \multicolumn{1}{c|}{5392}            & -                 & \multicolumn{1}{c|}{472}             & \multicolumn{1}{c|}{764}             & \multicolumn{1}{c|}{512}                          & \multicolumn{1}{l|}{1387}                & \multicolumn{1}{l|}{224}               & 2033                                \\ \hline
HASOC                                                       & \multicolumn{1}{c|}{1237}            & 4348              & \multicolumn{6}{c|}{-}                                                                                                                                                                                                                                    \\ \hline
AYR                                                       & \multicolumn{1}{c|}{393}             & 1246              & \multicolumn{1}{c|}{42}              & \multicolumn{1}{c|}{343}             & \multicolumn{4}{c|}{-}                                                                                                                                                      \\ \hline
AHSD                                                       & \multicolumn{1}{c|}{1363}            & 4088              & \multicolumn{6}{c|}{-}                                                                                                                                                                                                                                    \\ \hline
HTPO                                                       & \multicolumn{1}{c|}{351}             & 2647              & \multicolumn{6}{c|}{-}                                                                                                                                                                                                                                    \\ \hline
HSHP                                                       & \multicolumn{1}{c|}{1498}             &     426          & \multicolumn{1}{c|}{9}               & \multicolumn{1}{c|}{1489}            & \multicolumn{4}{c|}{-}                                                                                                                                                      \\ \hline
\textbf{Total}                                          & \multicolumn{1}{c|}{\textbf{27,579}} & \textbf{55,651}   & \multicolumn{1}{c|}{\textbf{8,216}}  & \multicolumn{1}{c|}{\textbf{7,938}}  & \multicolumn{1}{c|}{\textbf{1,170}}               & \multicolumn{1}{c|}{\textbf{1,438}}      & \multicolumn{1}{c|}{\textbf{709}}      & \multicolumn{1}{c|}{\textbf{2,186}} \\ \hline
\end{tabular}
}

\caption{Distribution of tweets gathered across hate speech datasets, including those where the target information is available (multiclass).}
\label{tab:data_distribution}
\end{table*}

        \subsubsection{Multiclass Setting}
        Having established our binary setting, we aggregated the available datasets aiming to construct a more detailed hate speech classification task. As an initial step, all available hate speech sub-classes present were considered.  However, this led to a very detailed but sparse hate taxonomy, with 44 different hate speech categories, but with only a few entries for some of the classes (e.g. "economic" category with only four tweets present). Aiming to create an easy-to-use and extendable data resource, several categories were grouped together. All classes related to ethnicity (e.g. Arab, Hispanic) or immigration were grouped under \textit{racism}, while religious categories (e.g. Muslim, Christian) were considered separately. Categories related to sexuality and sexual orientation (e.g. heterosexual, homosexual) were also grouped in one class, and tweets with topics regarding gender (men, women) constitute the \textit{sexism} class. Finally, all entries labelled as "not-hate" speech were also included. To keep our dataset relatively balanced we also ignored classes that constitute less than 1\% of the total hate speech data. Overall, the multiclass setting proposed consists of 7 classes: \textit{Racism, Sexism, Disability, Sexual orientation, Religion, Other}, and \textit{Not-Hate}. It is worth noting that tweets falling under the \textit{Other} class do not belong to any of the other five hate speech classes.

        \subsubsection{Statistics and Data Splits}
        \label{summarydata}
        In total, we collected 83,230 tweets, from 13 different datasets (Table \ref{tab:data_distribution}), of which only 33\% are classified as hate speech. This unified dataset may seem imbalanced but it is commonly assumed that only around 1\% of the content shared on social media contains hate speech \cite{pereira2019detecting}. When considering the multiclass setting, the hate speech percentage decreases even more with only 26\% of tweets labelled as a form of hate speech, with the \textit{religion} class being the least popular with only 709 entries.

        The data in both settings (binary \& multiclass) are divided into train and test sets using a stratified split to ensure class balance between the splits (Table \ref{tab:train_test_distributions}). In general, for each dataset present, we allocate 70\% as training data, 10\% as validation, and 20\% as test data. Exceptions to the aforementioned approach are datasets where the authors provide a preexisting data split which we use.

\begin{table}[ht]
\scalebox{0.91}{
\begin{tabular}{|l|rr|rr|}
\hline
\multicolumn{1}{|c|}{\multirow{2}{*}{\textbf{Dataset}}} & \multicolumn{2}{c|}{\textbf{train}}                                         & \multicolumn{2}{c|}{\textbf{test}}                                          \\ \cline{2-5} 
\multicolumn{1}{|c|}{}                                  & \multicolumn{1}{c|}{\textbf{not-hate}} & \multicolumn{1}{c|}{\textbf{hate}} & \multicolumn{1}{c|}{\textbf{not-hate}} & \multicolumn{1}{c|}{\textbf{hate}} \\ \hline
AHSD                                                    & \multicolumn{1}{r|}{3270}              & 1090                               & \multicolumn{1}{r|}{818}               & 273                                \\ \hline
AYR                                                     & \multicolumn{1}{r|}{996}               & 314                                & \multicolumn{1}{r|}{250}               & 79                                 \\ \hline
CMS                                                     & \multicolumn{1}{r|}{8688}              & 989                                & \multicolumn{1}{r|}{2173}              & 248                                \\ \hline
DEAP                                                    & \multicolumn{1}{r|}{84}                & 2981                               & \multicolumn{1}{r|}{21}                & 746                                \\ \hline
HASOC*                                                   & \multicolumn{1}{r|}{3489}              & 1113                               & \multicolumn{1}{r|}{859}               & 124                                \\ \hline
HSHP                                                    & \multicolumn{1}{r|}{341}               & 1197                               & \multicolumn{1}{r|}{85}                & 301                                \\ \hline
HTPO*                                                    & \multicolumn{1}{r|}{2106}              & 292                                & \multicolumn{1}{r|}{541}               & 59                                 \\ \hline
HatE*                                                 & \multicolumn{1}{r|}{5757}              & 4197                               & \multicolumn{1}{r|}{1607}              & 1106                               \\ \hline
HateX                                              & \multicolumn{1}{r|}{4542}              & 2050                               & \multicolumn{1}{r|}{1136}              & 512                                \\ \hline
LSC                                                     & \multicolumn{1}{r|}{1013}              & 711                                & \multicolumn{1}{r|}{254}               & 178                                \\ \hline
MHS                                                     & \multicolumn{1}{r|}{4058}              & 1988                               & \multicolumn{1}{r|}{1016}              & 497                                \\ \hline
Offense*                                              & \multicolumn{1}{r|}{10037}             & 913                               & \multicolumn{1}{r|}{2510}               & 229                                 \\ \hline
\textbf{All}                                              & \multicolumn{1}{r|}{\textbf{44,381}}             & \textbf{17,835}                               & \multicolumn{1}{r|}{\textbf{11,270}}               & \textbf{4,352}                                 \\ \hline
\end{tabular}
}
\caption{Binary class distribution in train and test splits of the unified hate speech datasets. * indicates datasets where preexisting train/test splits were available and retrieved.}
\label{tab:train_test_distributions}
\end{table}

\begin{table*}[ht]
\setlength{\tabcolsep}{2.8pt}
\begin{adjustbox}{width=\textwidth,center}
\begin{tabular}{lll|ccccccccccccc|l}
\hline
\multicolumn{1}{l|}{\multirow{2}{*}{\textbf{Model}}} & \multicolumn{2}{c|}{\textbf{Train}} & \multirow{2}{*}{\textbf{HatE}} & \multirow{2}{*}{\textbf{MHS}} & \multirow{2}{*}{\textbf{DEAP}} & \multirow{2}{*}{\textbf{CMS}} & \multirow{2}{*}{\textbf{Off.}} & \multirow{2}{*}{\textbf{HateX}} & \multirow{2}{*}{\textbf{LSC}} & \multirow{2}{*}{\textbf{HASOC}} & \multirow{2}{*}{\textbf{AYR}} & \multirow{2}{*}{\textbf{AHSD}} & \multirow{2}{*}{\textbf{HTPO}} & \multirow{2}{*}{\textbf{HSHP}} & \multirow{2}{*}{\textbf{AVG}} & \multirow{2}{*}{\textbf{Indep}} \\ \cline{2-3}
\multicolumn{1}{l|}{}                                & \textbf{Data}             & \textbf{Size}             &                                &                               &                                &                               &                                   &                                 &                               &                                 &                               &                                &                                &                                &                               &                                  \\ \hline
\multicolumn{1}{l|}{\multirow{3}{*}{BERTweet}}                        & All              & 58213            & 57.1                           & 87.7                          & \textbf{57.7}                  & \textbf{82.4}                 & 59.4                              & \textbf{75.1}                   & 61.5                          & 59.4                            & 85.5                          & \textbf{90.2}                  & 59.5                           & \textbf{65.4}                  & \textbf{70.1}                 & 61.0                             \\
\multicolumn{1}{l|}{}                                & All*            & 5290             & 51.1                           & 80.5                          & 53.7                           & 73.1                          & \textbf{60.8}                     & 67.3                            & 72.1                          & \textbf{63.9}                   & \textbf{85.6}                 & 85.4                           & 67.6                           & 62.1                           & 68.6                          & \textbf{69.2}                    \\
\multicolumn{1}{l|}{}                                & MHS              & 5291             & \textbf{65.5}                  & \textbf{89.3}                 & 13.3                           & 50.6                          & 53.2                              & 69.6                            & 58.8                          & 58.0                            & 66.8                          & 78.8                           & \textbf{67.7}                  & 28.6                           & 58.3                          & 58.6                             \\ \hline
\multicolumn{1}{l|}{\multirow{3}{*}{TimeLMs}}        & All              & 58213            & 54.2                           & \textbf{86.6}                 & \textbf{68.0}                  & \textbf{79.7}                 & \textbf{56.9}                     & \textbf{74.8}                   & 59.1                          & 63.2                            & 87.2                          & \textbf{89.4}                  & 65.2                           & \textbf{64.5}                  & \textbf{70.7}                 & 63.7                             \\
\multicolumn{1}{l|}{}                                & All*            & 1146             & 48.3                           & 74.9                          & 49.3                           & 69.3                          & 54.7                              & 59.7                            & \textbf{63.8}                 & \textbf{63.8}                   & 82.3                          & 79.9                           & 59.6                           & 63.0                           & 64.0                          & \textbf{70.6}                    \\
\multicolumn{1}{l|}{}                                & AYR              & 1147             & \textbf{61.0}                  & 71.4                          & 9.8                            & 63.5                          & 52.5                              & 56.3                            & 60.9                          & 63.6                            & \textbf{87.7}                 & 80.7                           & \textbf{66.8}                  & 57.9                           & 61.0                          & 59.3                             \\ \hline
\multicolumn{1}{l|}{\multirow{3}{*}{RoBERTa}}        & All              & 58213            & 52.3                           & \textbf{85.9}                 & \textbf{66.6}                  & \textbf{79.9}                 & \textbf{54.7}                     & \textbf{73.8}                   & 59.5                          & 60.8                            & \textbf{87.0}                 & \textbf{89.8}                  & 64.4                           & 61.4                           & \textbf{69.7}                 & 56.2                             \\
\multicolumn{1}{l|}{}                                & All*            & 1146             & \textbf{56.0}                  & 73.7                          & 53.2                           & 64.2                          & 53.0                              & 48.9                            & \textbf{70.2}                 & \textbf{65.8}                   & 74.3                          & 74.1                           & 58.9                           & 61.0                           & 62.8                          & \textbf{78.3}                    \\
\multicolumn{1}{l|}{}                                & AYR              & 1147             & 54.8                           & 63.8                          & 17.5                           & 69.8                          & 55.2                              & 50.1                            & 57.7                          & 63.4                            & 86.3                          & 81.9                           & \textbf{64.6}                  & 55.6                           & 60.1                          & 53.8                             \\ \hline
\multicolumn{1}{l|}{\multirow{3}{*}{BERT}}           & All              & 58213            & 52.3                           & \textbf{84.0}                 & 49.3                           & 79.7                          & \textbf{56.8}                     & \textbf{74.1}                   & 56.9                          & \textbf{60.9}                   & \textbf{85.2}                 & \textbf{89.6}                  & 60.5                           & \textbf{65.5}                  & \textbf{67.9}                 & 50.7                             \\
\multicolumn{1}{l|}{}                                & All*            & 2098             & 44.7                           & 75.0                          & 49.2                           & \textbf{66.1}                 & 55.9                              & 59.1                            & \textbf{63.5}                 & 60.5                            & 71.1                          & 74.1                           & 57.0                           & 60.5                           & 61.4                          & \textbf{60.9}                    \\
\multicolumn{1}{l|}{}                                & HTPO             & 2099             & 54.9                           & 77.5                          & 19.8                           & 52.1                          & 52.1                              & 58.6                            & 64.8                          & 55.9                            & 61.3                          & 78.1                           & \textbf{73.5}                  & 38.3                           & 57.2                          & 50.7                             \\ \hline
\multicolumn{1}{l|}{\multirow{3}{*}{SVM}}            & All              & 58213            & 50.6                           & 77.0                          & \textbf{61.6}                  & 66.1                          & 48.5                              & 71.2                            & 47.8                          & 48.9                            & \textbf{86.9}                 & \textbf{87.3}                  & 47.3                           & 54.9                           & 61.2                          & 46.7                             \\
\multicolumn{1}{l|}{}                                & All*            & 5290             & 44.5                           & 76.1                          & 55.7                           & \textbf{68.4}                 & \textbf{50.7}                     & 64.4                            & \textbf{57.0}                 & \textbf{56.2}                   & 81.0                          & 81.9                           & \textbf{52.7}                  & 57.4                           & \textbf{67.2}                 & \textbf{59.3}                    \\
\multicolumn{1}{l|}{}                                & MHS              & 5291             & \textbf{57.9}                  & \textbf{80.0}                 & 4.8                            & 48.3                          & 48.4                              & \textbf{67.2}                   & 46.4                          & 46.4                            & 47.8                          & 75.0                           & 50.1                           & 22.7                           & 47.7                          & 51.8                             \\ \hline
\multicolumn{3}{l|}{All hate baseline}                                                              & 29.0                           & 25.0                          & 49.0                           & 9.0                           & 8.0                               & 24.0                            & 29.0                          & 11.0                            & 19.0                          & 20.0                           & 9.0                            & 44.0                           & 23.0                          & 10.0                             \\ \hline
\end{tabular}
\end{adjustbox}
\caption{Macro-averaged F1 scores across all hate speech test sets and our manually annotated set (Indep). For each model, the table includes:  (1) the performance of the model trained on all the datasets (All); (2) the performance of the model when trained on a balanced sample of all datasets of the same size as the best single-dataset baseline (All*); and (3) the best overall performing model trained on a single dataset (BERTweet: \textit{MHS}, TimeLMs: \textit{AYR}, RoBERTa: \textit{AYR}, BERT: \textit{HTPO}, SVM: \textit{MHS}). The best result for each dataset and model is bolded.}
\label{tab:grouped_results}
\end{table*}

\section{Evaluation}

We present our main experimental results comparing various language models trained on single datasets and in the unified dataset presented in the previous section.

    \subsection{Experimental Setting}

    \paragraph{Models.} For our experiments we rely on four language models of a similar size, two of them being general-purposes and the other two specialized on social media: BERT-base \cite{DBLP:journals/corr/abs-1810-04805} and RoBERTa-base \cite{DBLP:journals/corr/abs-1907-11692} as general-purpose models; and BERTweet \cite{nguyen-etal-2020-bertweet} and TimeLMs-21 \cite{loureiro-etal-2022-timelms} as language models specialized on social media, and particularly Twitter. There is an important difference between BERTweet and TimeLMs-21: since BERTweet was trained from scratch, TimeLMs-21 used the RoBERTa-base checkpoint as initialization and then continued training on a Twitter corpus. An SVM classifier is also utilized as a baseline model. 
    
     \paragraph{Settings.} Aiming to investigate the effect of a larger and more diverse hate speech training corpus on various types of hate speech, we perform an evaluation on both the binary and multiclass settings described in Section \ref{unification}. Specifically, for the binary setting we fine-tune the models selected first on each individual dataset, and secondly while using the unified dataset created. For the multiclass setting, we considered the unified and the HateX dataset, which includes data for all classes. In total, we fine-tuned 54 different binary\footnote{MMHS dataset was used only for the training/evaluation of the unified dataset as it is lacking the \textit{not-hate} class} and 8 multiclass models.

     \paragraph{Training.} The implementations provided by Hugging Face \cite{wolf-etal-2020-transformers} are used to train and evaluate all language models, while we utilise Ray Tune \cite{liaw2018tune} along with HyperOpt \cite{bergstra2022hyperopt} and Adaptive Successive Halving \cite{li2018massively} for optimizing the learning rate, warmup steps, number of epochs, and batch size, hyper-parameteres of each model.\footnote{Optimal hyperparameters can be found in Table \ref{tab:hyperparameters} in the Appendix}

     \paragraph{Evaluation metrics.} The macro-averaged F1 score is reported and used to compare the performance of the different models. Macro-F1 is commonly used in similar tasks  \cite{basile-etal-2019-semeval, zampieri-etal-2020-semeval} as it provides a more concrete view on the performance of each model.
    
\begin{table*}[ht]
\scalebox{0.88}{
\begin{tabular}{|l|l|c|c|c|c|c|c|c||c|}
\hline
\textbf{model}            & \textbf{Train} & \textbf{sexism} & \textbf{racism} & \textbf{disability} & \textbf{\begin{tabular}[c]{@{}c@{}}sexual\\ orientation\end{tabular}} & \textbf{religion} & \textbf{other} & \textbf{not-hate} & \textbf{AVG} \\ \hline
\multirow{2}{*}{TimeLMs}  & All Datasets   & \textbf{72.2}  & \textbf{72.9}  & \textbf{74.2}      & \textbf{76.9}    & \textbf{52.6}    & \textbf{58.8} & \textbf{90.6}    & \textbf{71.6}     \\ \cline{2-10} 
                          & HateX     & 52.1           & 16.5           & 0           & 58.8   & 31.8        & 5.8           & 86.0           & 35.9              \\ \hline
\multirow{2}{*}{BERTweet} & All Datasets   & \textbf{73.1}  & \textbf{72.5}  & \textbf{74.1}       & \textbf{77.6} & \textbf{48.6}    & \textbf{59.3} & \textbf{90.9}    & \textbf{70.9}     \\ \cline{2-10} 
                          & HateX     & 47.8        & 6.8         & 0           & 43.9  & 0     & 0        & 85.5            & 26.3              \\ \hline
\multirow{2}{*}{RoBERTa}  & All Datasets   & \textbf{70.4}  & \textbf{72.4}  & \textbf{73.9}      & \textbf{76.5}   & \textbf{47.3}    & \textbf{55.5} & \textbf{90.3}    & \textbf{69.5}     \\ \cline{2-10} 
                          & HateX     & 50.5          & 16.3           & 0              & 67.9   & 29.1             & 7.7           & 85.5             & 36.3              \\ \hline
\multirow{2}{*}{BERT}     & All   & \textbf{68.9}  & \textbf{66.3}  & \textbf{75.5}      & \textbf{69.3}  & \textbf{40.3}     & \textbf{54.9} & \textbf{93.3}    & \textbf{66.9}     \\ \cline{2-10} 
                          & HateX     & 40.4           & 16.0           & 0                   & 66.2  & 15.9         & 0              & 85.4             & 32.0              \\ \hline
\multirow{2}{*}{SVM}     & All & \textbf{62.7} & \textbf{67.0} & \textbf{71.5} & \textbf{70.5} & 4.1 & \textbf{49.0} & \textbf{59.11} & \textbf{81.9} \\ \cline{2-10} 
                          & HateX  & 20.1 & 6.0 & 0 & 54.9 & \textbf{6.8} & 0 & 84.5 & 24.6 \\ \hline                          
\multicolumn{2}{|l|}{Baseline (most frequent)}                     & 0               & 0               & 0                   & 0            & 0                 & 0              & 84.0                & 12.0                 \\ \hline
\end{tabular}
}
\caption{F1 score for each class in the multiclass setting when trained on all the datasets (All) and when trained only with HateX. Macro-average F1 (AVG) is also reported.}
\label{tab:multiclass_results}
\end{table*}
    
     \subsection{Datasets}
     \label{subsec:datasets}

     For training and evaluation, we use the splits described in Section \ref{summarydata}. As described above, for each language model we trained on each dataset training set independently, and in the combination of all dataset-specific training sets. The results on the combination of all datasets are averaged across each dataset-specific test set (AVG), i.e., each dataset is given the same weight irrespective of its size. In addition to the datasets presented in Section \ref{hatedatasets}, we constructed an independent test set (Indep) to test the robustness of models outside existing datasets.

     \paragraph{Independent test set (Indep).} This dataset was built by utilising a set of keywords related to the \textit{International Women's Day} and \textit{International Day Against Homophobia, Transphobia and Biphobia} and extracting tweets from the respected days of 2022. Then, these tweets were manually annotated by an expert. In total 200 tweets were annotated as hateful, not-hateful, or as "NA" in cases where the annotator was not sure whether a tweet contained hate speech or not. The \textit{Indep} test set consists of 151 non-hate and 20 hate tweets and due to its nature (specific content \& expert annotation) can be leveraged to perform a targeted evaluation on models trained on similar and unrelated data. While we acknowledge the limitations of the \textit{Indep} test set (i.e., relative small number of tweets and only one annotator present), our aim is to use these tweets, collected using relatively simple guidelines\footnote{Annotator guidelines are available in Appendix \ref{sec:annotation_guidelines}.}, to test the overall generalisation ability of our models and how it aligns to what people think of hate speech. 

\subsection{Results}

\subsubsection{Binary Setting}
Table \ref{tab:grouped_results} displays the macro-F1 scores achieved by the models across all test sets when fine-tuned: (1) on all available datasets (\textit{All}), (2) on the best overall performing model trained on a single dataset, and (3) on a balanced sample of the unified dataset of the same data size as (2). When looking at the average performance of (1) and (2), it is clear that when utilising the combined data, all models perform considerably better overall. 
This increased performance may not be achieved across all the datasets tested, but it does provide evidence that the relatively limited scope of the individual datasets hinder the potential capabilities of our models. An even bigger constrast is observed when considering the performance difference on the \textit{DEAP} subset, which deals with a less common type of hate speech (prejudice towards Asian people),  where even the best performing single dataset model achieves barely 19.79\% F1 compared to the worst combined classifier with 49.27\% F1 (BERT All / BERT \textit{HTPO}).

To further explore the importance of the size and diversity of the training data we train and evaluate our models in an additional settings. Considering the sample size of the best performing dataset for each model, an equally sized training set is extracted from all available data while enforcing a balanced distribution between hate and not-hate tweets (\textit{All*}). Finally, we make sure to sample proportionally across the available datasets.
The results (Table \ref{tab:grouped_results}) reveal the significance that a diverse dataset has in the models' performance. All models tested perform on average better when trained on the newly created subsets (\textit{All*}) when compared to the respective models trained only on the best performing individual dataset. Interestingly, this setting also achieves the best overall scores on the \textit{Indep.} set, which reinforces the importance of balancing the data. 
Nonetheless, all the transformers models still achieve their best score when trained on all the combined datasets (\textit{All}) which suggests that even for these models, the amount of available training data remains an important factor of their performance.

\begin{figure*}[!ht]
\centering
\begin{adjustbox}{width=\textwidth,center}
\includegraphics{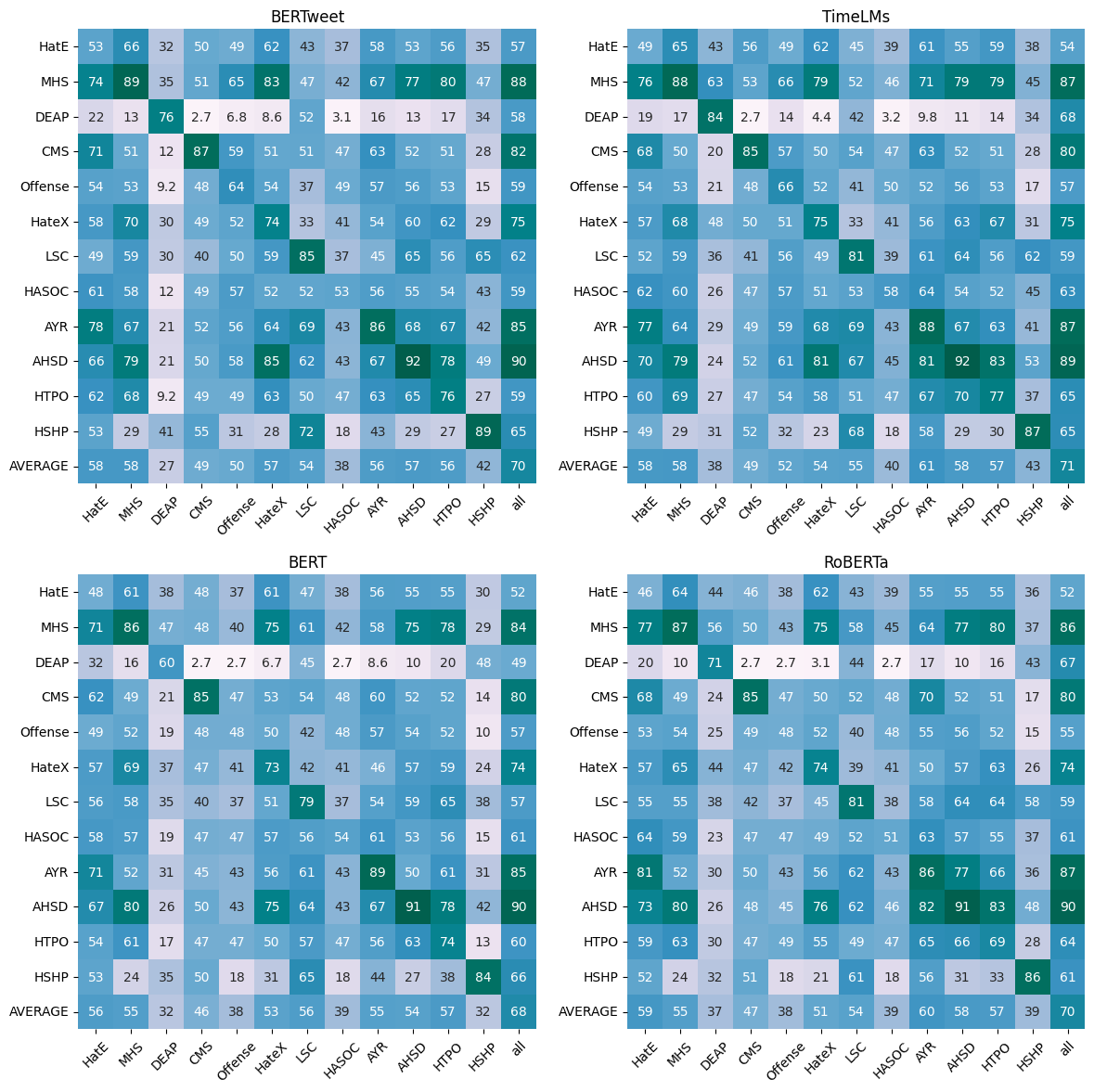}
\end{adjustbox}
\caption{Macro-averaged F1 score for each dataset/model combination. The X axis indicates on which dataset the model was trained while the Y axis indicates the test set used to evaluate it. \textit{AVERAGE} indicates the result by averaging across all datasets, and \textit{all} represents the aggregated training set including all datasets.}
\label{fig:f1_heatmap}
\end{figure*}

\subsubsection{Multiclass Setting}
Similarly to our binary setting, utilising the combined datasets in the multiclass setting enhances the models' performance. As can be observed from Table \ref{tab:multiclass_results}, all the models struggle to function at a satisfactory degree when trained on the \textit{HateX} subset only. In particular, when looking at the "disability" class, none of the models manage to classify any of the entries correctly. This occurs even though "disability" entries exist in the \textit{HateX} training subset, albeit in a limited number (21). This behaviour suggests that even when information about a class is available in the training data, language models may fail to distinguish and utilise it. Imbalanced datasets are a common challenge in machine learning applications. This issue is also present in hate speech, in this case exacerbated given the nature of the problem (including a potential big overlap of features between classes) and the lack of resources available. 

\label{resultsmulticlass}

\section{Analysis}

In this section, we dissect the results presented in the previous section by performing a cross-dataset comparison and a qualitative error analysis.

\subsection{Cross-dataset Analysis}

Figure \ref{fig:f1_heatmap} presents a cross-dataset comparison of the language models used for the evaluation. The heatmap presents the results of the models fine-tuned and tested for all dataset pair combinations. All models evaluated tend to perform better when they are trained and tested on specific subsets (left diagonal line on the heat-maps). Even when we evaluate models on similar subsets, they tend to display a deterioration in performance. For example both \textit{CMS} and \textit{AYR} datasets deal with sexism but the models trained only on \textit{CMS} perform poorly when evaluated on \textit{AYR} (e.g. BERTweet-CSM achieves 87\% F1 on \textit{CSM}, but only 52\% on \textit{AYR}). Finally, it is observable again that the models trained on the combined datasets (column "all") display the best overall performance and attain consistently high results in each individual test set. When analysing the difficulty of each individual dataset when used as a test set, DEAP is clearly the most challenging one overall.  This may be due to the scope of the dataset, dealing with East Asian Prejudice during the COVID-19 pandemic, which is probably not well captured in the rest of the datasets. When used as training sets, none of the individual datasets is widely generalisable, with the results of the model fine-tuned on them being over 10 points lower than when fine-tuned on the unified dataset in all cases.

\subsection{Qualitative Error Analysis}
Aiming to better understand the models' results we perform a qualitative analysis focusing on entries miss-classified by our best performing model, \textit{TimeLMs-All}.

\paragraph{Multiclass.} When considering the multiclass setting, common errors are tweets that have been labelled as hateful,   e.g. "U right, probably some old n*gga named Clyde" is labelled as \textit{racism} and "@user @user she not a historian a jihadi is the correct term" as \textit{religion}, but the model classifies them as \textit{not-hate}. However, depending on the context and without having access to additional information (author/target of the tweet) these entries may not actually be hateful. 

It is also interesting to note the limitations that arise when training only on a single dataset, particularly if the data collection is done by utilising specific keywords. For example the tweets "Lana i love you b*tch. Put that flag back up h*e \#lustfoflife" and "happy birthday b*tch, hope you have a great one h*e! @user" are correctly classified as \textit{not-hate} by \textit{TimeLMs-All} but are miss-classified as \textit{sexism} by \textit{TimeLMs-HateX}, despite \textit{sexism} being present in the \textit{HateX} dataset.

\paragraph{Binary}
In the binary setting, the model seems to struggle with entries such as "Meanwhile in Spain..\#stopimmigration" and "This is outrageous. Congress should be fired on the spot. \#BuildThatWall \#stopwastingmytaxdollars" where both entries are classified as \textit{hate} but are labelled as \textit{not-hate}. Similarly to the previous case, the classification of such tweets without additional context is a difficult task. While these tweets have hateful undertones, they may not be necessarily hate speech without considering them in their broader context.

Finally, when looking at the classification errors of \textit{TimeLMs-AYR} (trained only on sexist and racist tweets) the need of diverse training data becomes apparent. For example, \textit{TimeLM-AYR} fails to classify as hate speech the tweets "@user that r*tarded guy should not be a reporter" and "I'm going to sell my iPhone and both my Macs, I don't support f*ggots." as hate speech in contrast to \textit{TimeLMs-All} which classifies the tweets correctly as hateful.

\section{Conclusion}

In this paper, we presented a large-scale analysis of hate speech detection systems based on language models. In particular, our goal was to show the divergences across datasets and the importance of having access to a diverse and complete training set. Our results show how the combination of datasets make for a robust model performing competitively across all datasets. This is not a surprising finding given the size of the corresponding training sets, but the considerable gap (e.g. 70.7\% to 61.0\% in Macro-F1 for the best TimeLMs-21 performing model) shows that models trained on single datasets have considerable room for improvement. Moreover, even when controlling for data size, a model trained on a diverse set instead of a single dataset leads to better overall results.


As future work, we are planning to extend this analysis beyond English, in the line of previous multilingual approaches \cite{ousidhoum-etal-2019-multilingual,chiril-etal-2019-multilingual,bigoulaeva-etal-2021-cross}, and masked language models by including, among others, generative and instruction-tuning language models. In addition to the extensive binary-level evaluation, recognising the target group is a challenging area of research. While in Section \ref{resultsmulticlass}, we provided some encouraging results, the results could be expanded with a unified taxonomy.

\section{Ethics Statement}
Our work aims to contribute and extend research regarding hate speech detection in social media and particular in Twitter. We believe that our efforts to contribute on the ongoing concerns around the status of hate speech on social medial.

We acknowledge the importance of the ACM Code of Ethics, and are committed on following it's guidelines. Our current work, uses either publicly available tweets under open licence and does not infringe any of the rules of Twitter's API. Moreover, given that our task includes user generated content we are committed to respect the privacy of the users, by replacing each user mention in the texts with a placeholder.

\section{Limitations}
In this paper, we have focused on existing datasets and a unification stemming from their features. The decisions taken to this unification, particularly in the selection of dataset and target groups, may influence the results of the paper. 

We have focused on social media (particularly Twitter) and on the English language. While there has been extensive work on this medium and language, the conclusions that we can take from this study can be limiting, as the detection of hate speech involves other areas, domains and languages. In general, we studied a particular aspect of hate speech detection which may or not be generalizable.

Finally, due to computational limitations, all our experiments are based on base-sized language models. It is likely that larger models, while exhibiting similar behaviours, would lead to higher results overall.

\section{Acknowledgements}

The authors are supported by a UKRI Future Leaders Fellowship. They also acknowledge the collaboration with the Spanish National Office Against Hate Crimes and the support of the EU Citizens, Equality, Rights and Values (CERV) programme. However, the authors have the exclusive responsibility for the contents of this publication. Finally, the authors thank Nina White for her help annotating the independent test set.

\bibliography{anthology,custom}
\bibliographystyle{acl_natbib}

\appendix

\section{Annotation Guidelines}
\label{sec:annotation_guidelines}

In the following we present the guidelines provided to the annotator for the independent test set (Section \ref{subsec:datasets}).

A tweet is:

\begin{itemize}
    \item labelled as "1" ("hate speech") if it contains any “discriminatory” (biased, bigoted or intolerant) or “pejorative” (prejudiced, contemptuous or demeaning) speech towards individuals or group of people.
    \item labelled as "0" ("not-hate-speech") if it does not contain hate speech as defined above.
    \item labelled "NA" if the coder is not sure whether the tweet contains hate speech or not.
\end{itemize}

The annotation should be based only on the text content of the tweet. This means that the coder should not follow any URL/media links if present.

\section{Hyperparameter Tuning}

Table \ref{tab:hyperparameters} lists the best hyperparameters for each of the models used in the evaluation.

\begin{table}[ht]
\centering
\scalebox{0.7}{
\begin{tabular}{llrrp{10mm}p{15mm}}
model & setting & learning rate & epochs & batch \newline size & warm-up\newline steps \\ \hline
TimeLMs & binary & 1.5857E-05 & 2 & 16 & 50 \\ 
BERTweet & binary & 1.4608E-05 & 2 & 4 & 100 \\ 
BERT & binary & 1.7882E-05	 & 2 & 4 & 10 \\ 
RoBERTa & binary & 1.0377E-05 & 2 & 4 & 50 \\ 
TimeLMs & multiclass &1.9100E-05	& 3 & 16 & 100 \\ 
BERTweet & multiclass & 9.0295E-06	 & 3 & 4 & 10 \\ 
BERT & multiclass & 8.1260E-06	 & 4 & 8 & 100 \\ 
RoBERTa & multiclass & 8.1260E-06	 & 4 & 16 & 10 \\ 
\end{tabular}}
\caption{Best hyper-parameters for models trained on the combined datasets for the binary and multiclass settings.}
\label{tab:hyperparameters}
\end{table}

\end{document}